\documentclass[11pt,letterpaper]{article}
\usepackage[T1]{fontenc}
\usepackage[utf8]{inputenc}
\usepackage[margin=0.9in]{geometry}
\usepackage{newtxtext,newtxmath}
\usepackage{amsmath}
\usepackage{microtype}
\usepackage{booktabs,tabularx,array}
\usepackage{graphicx}
\usepackage{xcolor}
\usepackage{tikz}
\usetikzlibrary{positioning,arrows.meta,calc}
\usepackage{caption}
\usepackage{enumitem}
\usepackage[numbers,sort&compress]{natbib}
\usepackage{xurl}
\usepackage{titlesec,needspace,float}
\usepackage{fancyvrb}
\usepackage[section]{placeins}
\floatstyle{ruled}
\newfloat{algorithm}{tbp}{loa}
\floatname{algorithm}{Algorithm}
\usepackage{hyperref}
\definecolor{linkblue}{RGB}{0,66,166}
\hypersetup{colorlinks=true,linkcolor=linkblue,citecolor=linkblue,urlcolor=linkblue,
 pdftitle={Evolutionary Ensemble Search: Council-Guided Program Evolution with Persistent Memory},
 pdfauthor={Juan P. Madrigal-Cianci and Eshan Chordia},
 pdfsubject={Executable algorithm search and an MLE-bench Lite development study},
 pdfkeywords={evolutionary search, machine-learning engineering, program mutation, multi-agent systems, automated machine learning}}
\titleformat{\section}{\large\bfseries}{\thesection}{.65em}{}
\titleformat{\subsection}{\normalsize\bfseries}{\thesubsection}{.65em}{}
\titlespacing*{\section}{0pt}{12pt}{6pt}
\titlespacing*{\subsection}{0pt}{8pt}{4pt}
\titlespacing*{\paragraph}{0pt}{5pt}{.55em}
\setlist{nosep,leftmargin=1.4em}
\newcolumntype{Y}{>{\raggedright\arraybackslash}X}
\newcommand{\EES}{\textsc{EES}}

\newcommand{\repofile}[2]{\href{https://github.com/impulse-ai/mlebench-medals/blob/51c39605da4e0347b416c44a6c13086200959d45/#1}{#2}}
\begin{document}
\begin{center}
{\LARGE\bfseries Evolutionary Ensemble Search}\\[5pt]
{\Large\bfseries Council-Guided Program Evolution with Persistent Memory}\\[15pt]
{\large Juan P. Madrigal-Cianci\qquad Eshan Chordia}\\[5pt]
Impulse AI\\[3pt]
\href{mailto:juan@impulselabs.ai}{\texttt{juan@impulselabs.ai}}\qquad
\href{mailto:eshan@impulselabs.ai}{\texttt{eshan@impulselabs.ai}}\\[6pt]
Technical Report --- September 2026
\end{center}
\vspace{3pt}
\begin{abstract}
Evolutionary Ensemble Search (EES) constructs machine-learning procedures through expert-guided program evolution. A role-specialized council turns task evidence and experimental results into structured search directions. An orchestrator allocates these directions to execution specialists and an evolutionary engine. The engine selects measured parents, diagnoses their errors, and produces descendants through code mutation, structured pipeline edits, and crossover. Each child must execute and acquire its own validation evidence. Population archives retain useful alternatives, while compatible predictions compete in a validation-gated ensemble stage. Search adapts through parent-relative operator credit, session memory, and problem-indexed lessons retrieved across runs. We specify these mechanisms, distinguish their execution profiles, and define the contracts required to compare candidates as their computations change. A public MLE-bench Lite development ledger records medal-threshold artifacts on 19 of 22 tasks (86.36\%), with best outcomes of 11 gold, five silver, and three bronze. The procedures span text, images, tables, audio, scientific geometry, and deterministic transformations. The campaign includes grade feedback between runs, external-source routes, and mixed confirmation procedures; its aggregate is an achieved development result, not a blind autonomous-agent success rate. The report contributes a concrete architecture for cumulative executable search and a versioned account of its cross-modal development outcomes.
\end{abstract}
\begin{center}\small
Public results and reproduction records: \href{https://github.com/impulse-ai/mlebench-medals/tree/51c39605da4e0347b416c44a6c13086200959d45}{\texttt{github.com/impulse-ai/mlebench-medals}}.
\end{center}

\section{Introduction}\label{sec:intro}
Machine-learning engineering involves choices about the computation itself: which data to use, how to represent them, what procedure to fit, and how to construct the final prediction. Search-based agents address these choices through code generation, targeted refinement, and repeated execution~\citep{nam2025,chen2026}. An effective search must also preserve alternatives, connect observed errors to subsequent edits, and retain experience that remains useful after the current experiment ends.

\EES{} addresses this problem with a council--orchestrator--evolution architecture. Council members examine the same experimental state through different roles: discovery, methodology, memory, modeling, and implementation review. They return complementary structured fields rather than vote on a final answer. The orchestrator turns their output into bounded experiments. Evolution then acts on executable candidate procedures: it selects parents using measured performance, mutates their code or pipeline structure, evaluates the children, and retains candidates for further reproduction or predictive combination.

Mutation is the principal transition between evaluated procedures. A parent with systematic residual error may motivate a feature transformation; a train--validation gap may motivate a different regularizer; a weak component may be replaced while the rest of the program is inherited. These are testable interventions, not accepted recommendations. The child has no fitness until it executes under the comparison protocol. This separation between proposing a change and measuring its outcome allows language-model reasoning to guide search without determining the result.

The architecture retains experience at three timescales. Operator rewards change the distribution of local mutations. Session memory carries findings and failed approaches into subsequent council reviews. A persistent library associates task fingerprints with lessons, strategy statistics, and adaptable candidates. These mechanisms implement operational self-learning: experience changes the search policy without updating the underlying language model's weights.

The contributions are:
\begin{enumerate}[itemsep=3pt,topsep=3pt]
\item A council-guided evolutionary architecture with explicit parent-conditioned mutation, crossover, screening, archive retention, and prediction-level recombination.
\item A formal account of candidate inheritance, validation comparability, operator credit, and cross-run memory, with the differences between EES execution profiles made explicit.
\item A versioned analysis of the MLE-bench Lite development campaign, including all task scores, confirmation identities, representative procedures, and the scope of reproducibility.
\end{enumerate}

Here, \emph{algorithmic discovery} denotes search over executable procedures and their computational structure. \emph{Scientific novelty} additionally requires comparison with prior methods. The Lite study measures achieved solution breadth and performance, not the contribution of each search component.

\section{Related Work}\label{sec:related}
\paragraph{Machine-learning engineering agents.}
MLE-bench evaluates the completion of Kaggle-derived ML tasks using local graders and historical human medal thresholds~\citep{chan2024}. MLE-STAR retrieves candidate approaches, uses ablations to identify influential code blocks, targets those blocks for refinement, and constructs ensembles~\citep{nam2025}. MARS combines cost-constrained Monte Carlo tree search, modular implementation, and comparative reflective memory~\citep{chen2026}. EES shares their execution-driven approach. Its organizing mechanism is the interaction between structured expert review, a persistent evolutionary population, parent-relative operator feedback, and cross-run retrieval. 

\paragraph{Evolutionary program search.}
Genetic programming searches executable structures through selection and variation~\citep{koza1992}. FunSearch combines language-model program proposals with evaluation and an evolutionary program database~\citep{romera2024}; AlphaEvolve extends evaluator-guided code evolution to mathematical and computational problems~\citep{novikov2025}. EES applies this principle to complete ML procedures. Its variation interface accepts both typed configuration changes and diagnostic code edits, while the evaluation interface retains the data, prediction, and validation semantics needed to compare heterogeneous candidates.

\paragraph{AutoML and predictive ensembles.}
Pipeline search, meta-learning, and reuse of previously evaluated models are established AutoML techniques~\citep{feurer2015,hutter2019}. Ensemble selection can exploit a library of heterogeneous predictors by optimizing their combination on validation data~\citep{caruana2004}. EES incorporates these capabilities rather than treating them as new in isolation. It also admits generated code, deterministic transformations, scientific feature construction, and relationship-based procedures. Predictive recombination is distinct from program crossover: it combines outputs of fitted candidates rather than creating a new training program from their components.

\paragraph{Reflective memory.}
Reflexion uses feedback-derived textual memory to influence subsequent trials without updating model weights~\citep{shinn2023}. MARS derives comparative lessons from differences between search branches~\citep{chen2026}. EES separates this reflective state from numerical operator credit and from task-indexed cross-run records. This distinction identifies where experience changes the search: the next council context, the next mutation distribution, or the initialization of a future task.

\section{Search Objects and Measurement Contracts}\label{sec:formulation}
\subsection{Task and executable candidate}
A task is $T=(D_{\mathrm{tr}},D_{\mathrm{te}},m,C,B)$, where $D_{\mathrm{tr}}$ contains available training examples, $D_{\mathrm{te}}$ contains test inputs, $m$ is the metric, $C$ is the output contract, and $B$ gives resource limits. The contract identifies sample rows, targets, shapes, and prediction semantics. Any additional data source has an explicit access status; source availability and permission are distinct from predictive usefulness.

An executable candidate is
\begin{equation}
 x=(u_x,z_x),\qquad u_x=(d_x,f_x,h_x,a_x,p_x,q_x),
 \label{eq:candidate}
\end{equation}
where $d_x$ specifies data access and training-side partitioning, $f_x$ the representation or feature program, $h_x$ the model or algorithmic structure, $a_x$ the fitting procedure, $p_x$ inference, and $q_x$ post-processing or ensemble topology. Hyperparameters and random seeds are fields of these components. The record $z_x$ stores ancestry, execution status, diagnostics, measurements, and artifact identities. A candidate can be materialized from a structured pipeline or from source code.

The mutable specification is separate from the comparison protocol $\kappa$. That protocol defines evaluation rows, target semantics, metric computation, and the information available during fitting. EES may explore a different training-side resampling scheme, but its fitness is comparable to a parent's only under a common $\kappa$. Changing the comparison split does not create a valid improvement by itself: both procedures must be reevaluated on a shared protocol before their scores are ranked together.

\subsection{Execution, eligibility, and cost}
The execution interface maps a candidate specification to predictions and a result record:
\begin{equation}
 \operatorname{Eval}_{\kappa}(u_x;D,B_x)
 \longrightarrow
 (\widehat Y_x^{\mathrm{val}},\widehat Y_x^{\mathrm{te}},Q_\kappa(x),z_x).
 \label{eq:evaluation}
\end{equation}
$D$ denotes the admitted inputs, $B_x$ is the allocated execution limit, and $Q_\kappa$ orients the task metric so that larger values are better. The score is defined only when execution produces usable, comparable evidence. A timeout, exception, or invalid artifact is a typed failure, not an arbitrary low fitness. Prediction records carry row identifiers, target order, class semantics, and validation provenance.

Let $\mathcal E_g^\kappa$ be the eligible evaluated candidates available at generation $g$. For a nonempty pool, the incumbent is
\begin{equation}
 x_g^\star\in\operatorname*{arg\,max}_{x\in\mathcal E_g^\kappa}Q_\kappa(x).
 \label{eq:incumbent}
\end{equation}
A sample-shaped fallback can satisfy $C$ without belonging to this validation-ranked pool. Equation~\eqref{eq:incumbent} selects among completed experiments.

Search consumes resources beyond training the selected candidate. For additive units such as device-hours, token expenditure, or monetary cost,
\begin{equation}
 C_{\mathrm{search}}=C_{\mathrm{profile/probe}}+C_{\mathrm{proposals}}
 +C_{\mathrm{execution}}+C_{\mathrm{ensemble}}+C_{\mathrm{control/memory}}.
 \label{eq:cost}
\end{equation}
Failed work is included. Parallel resource use is distinct from elapsed wall time. Job deadlines and caps on generations, population size, parents, and offspring bound execution.

\Needspace{6\baselineskip}
\subsection{Population state and implementation profiles}\label{sec:profiles}
The common search state is
\begin{equation}
 \mathcal S_g=(P_g,A_g,K_g,\pi_g,R_g),
 \label{eq:state}
\end{equation}
where $P_g$ is the active evaluated population, $A_g$ an archive of reusable candidates, $K_g$ session memory and retrieved context, $\pi_g$ the mutation-operator distribution, and $R_g$ the remaining budget. An execution log is maintained separately from $A_g$: failed candidates remain in the log even when they are ineligible for the population or Pareto archive.

EES provides three execution profiles (Table~\ref{tab:profiles}). They share candidate, evaluation, and lineage interfaces but do not use identical search policies. The council is the task-facing control layer; it can direct specialist work or submit an evolutionary job. Section~\ref{sec:protocol} states which aspects the campaign record can substantiate.

\begin{table}[htbp]
\centering\small
\caption{EES execution profiles. The nested profile supplies the detailed adaptive search policy in Sections~\ref{sec:mutation}--\ref{sec:memory}.}\label{tab:profiles}
\begin{tabularx}{\linewidth}{@{}p{1.00in}YY@{}}\toprule
\textbf{Profile}&\textbf{Search representation}&\textbf{Selection and feedback}\\\midrule
Multi-island&Generated training code in general, tree-ensemble, and neural populations.&Within-island tournament selection, code mutation and crossover, and Pareto consolidation.\\\addlinespace[4pt]
Nested&Structured pipelines and optional code candidates.&Diagnostic interventions, fitness--offspring parent sampling, operator credit, archive reuse, and cross-run lessons.\\\addlinespace[4pt]
Benchmark&Registered modality operators and bounded configuration variants.&Comparable-validation frontier, hypothesis probes, prediction alignment, and terminal benchmark grading.\\\bottomrule
\end{tabularx}
\end{table}

\section{Council-Guided Evolutionary Search}\label{sec:architecture}
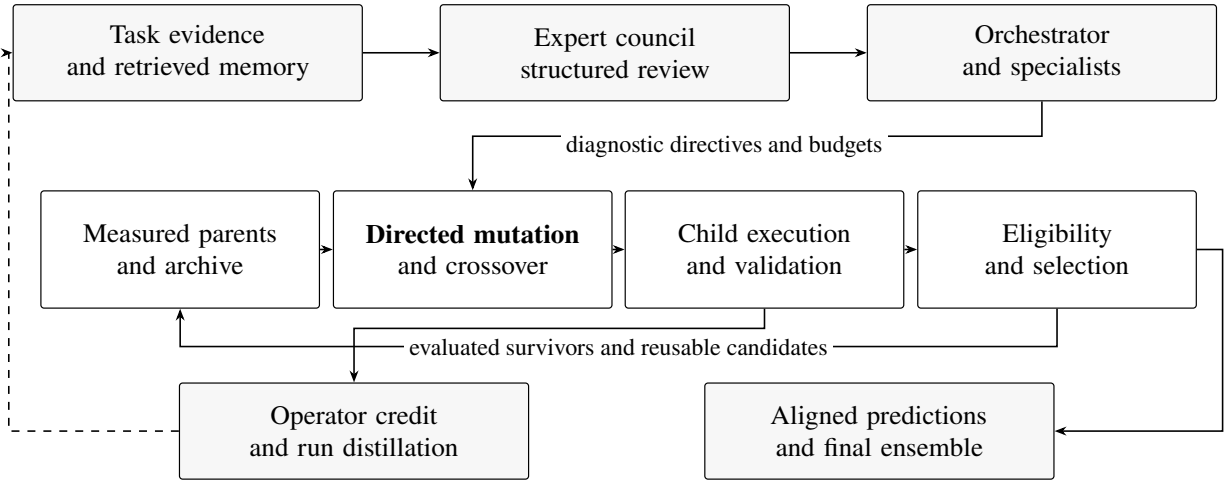
\begin{figure}[htbp]
\centering
\begin{tikzpicture}[
 big/.style={draw,rounded corners=1.2pt,line width=.45pt,fill=black!3,align=center,text width=1.68in,minimum height=.50in,inner sep=5pt,font=\small},
 smallbox/.style={draw,rounded corners=1.2pt,line width=.45pt,align=center,text width=1.34in,minimum height=.61in,inner sep=4pt,font=\small},
 flow/.style={-{Stealth[length=4.5pt]},line width=.6pt},
 note/.style={font=\footnotesize,fill=white,inner sep=2pt}]
\node[big] (memory) at (0,0) {Task evidence\\and retrieved memory};
\node[big] (council) at (5.65,0) {Expert council\\structured review};
\node[big] (orch) at (11.3,0) {Orchestrator\\and specialists};
\draw[flow] (memory)--(council); \draw[flow] (council)--(orch);
\node[smallbox] (parents) at (-.10,-2.6) {Measured parents\\and archive};
\node[smallbox] (mut) at (3.77,-2.6) {\textbf{Directed mutation}\\and crossover};
\node[smallbox] (exec) at (7.63,-2.6) {Child execution\\and validation};
\node[smallbox] (sel) at (11.5,-2.6) {Eligibility\\and selection};
\draw[flow] (parents)--(mut);\draw[flow] (mut)--(exec);\draw[flow] (exec)--(sel);
\draw[flow] (orch.south)--++(0,-.46)-| (mut.north);
\node[note] at (7.1,-1.24) {diagnostic directives and budgets};
\draw[flow] (sel.south)--++(0,-.50)-| (parents.south);
\node[note] at (5.7,-3.89) {evaluated survivors and reusable candidates};
\node[big] (credit) at (2.20,-5.0) {Operator credit\\and run distillation};
\node[big] (final) at (9.15,-5.0) {Aligned predictions\\and final ensemble};
\draw[flow] (exec.south)--++(0,-.26)-| (credit.north);
\draw[flow] (sel.east)--++(.34,0)|- (final.east);
\draw[flow,dashed] (credit.west)--++(-2.25,0)|- (memory.west);
\end{tikzpicture}
\caption{Council-guided executable search. The inner loop selects measured parents, constructs edits, executes children, and updates the population. Numerical operator feedback and distilled run records influence subsequent proposals. Predictive ensembling occurs only after alignment and validation checks; the external benchmark grade is outside this loop.}\label{fig:architecture}
\end{figure}

Figure~\ref{fig:architecture} shows the control and execution interfaces. The council proposes and critiques directions; the orchestrator assigns work; variation creates child specifications; execution supplies the evidence used by selection. 

\subsection{Council, orchestration, and initial evidence}\label{sec:council}
At council round $r$, expert $e$ receives task $T$, compact memory $K_r$, and recent measurements $V_r$. It emits an expert-specific record $u_{e,r}=\Phi_e(T,K_r,V_r)$. The runner executes experts concurrently and merges their complementary fields into
\begin{equation}
 U_r=\mathcal M\bigl(\{u_{e,r}:e\in\mathcal C_r\}\bigr),
 \label{eq:council}
\end{equation}
where $\mathcal C_r$ is the active role set. Table~\ref{tab:council} lists the roles. A model recommendation can coexist with a leakage warning or a record that the approach has already failed. Selection remains governed by executable evaluations.

The orchestrator consults the council before substantive execution and after material results. It assigns data profiling, approach research, domain review, implementation, training, inference, or synthesis to specialists. Council outputs remain advisory: specialists and registered tools perform the artifact-changing operations. Council rounds and evolutionary generations have different cadences; a review can redirect multiple subsequent generations.

\begin{table}[htbp]
\centering\small
\caption{Role-specialized council outputs. The first three roles are always active; the final two occupy rotating slots.}\label{tab:council}
\begin{tabularx}{\linewidth}{@{}p{1.34in}Y@{}}\toprule
\textbf{Expert}&\textbf{Contribution to the search state}\\\midrule
Discovery Driver&Next experiment, urgency, executable proposal, and a low-cost opportunity.\\
Methodology Advisor&Problem formulation, methodological fit, confidence, and checkpoint advice.\\
Memory Keeper&Established findings, attempted approaches, failed routes, and unresolved questions.\\
ML Strategist&Model family, feature intervention, validation design, and convergence assessment.\\
Code Reviewer&Implementation defects, statistical errors, leakage risks, and corrective actions.\\\bottomrule
\end{tabularx}
\end{table}

Initialization inventories tables, media, archives, auxiliary scientific files, and the sample submission. The resulting profile records identifiers, targets, modality, metric direction, sample counts, and file topology. A hypothesis links an observation to a candidate route and specifies prerequisites, expected artifacts, promotion criteria, and probe and execution budgets. Probes test whether the route is applicable: geometry requires auxiliary-file coverage; relationship matching requires identifiable correspondences; provenance matching requires measured precision and coverage. A promoted route initializes a candidate or constrains a mutation. Retrieved lessons supply priors, not exemptions from the task's evaluation and access requirements.

\subsection{Directed mutation: from parent evidence to child program}\label{sec:mutation}
Mutation starts from an evaluated parent, not only from the task description. For parent $x$, the diagnostic record $\eta_x$ can include validation residuals, errors by data segment, train--validation gaps, fold variability, feature gaps, and prediction correlations with other candidates. The meso-level critic interprets this record; an architect converts the diagnosis into an edit directive and, where appropriate, a bias over variation operators.

For operator $o$, mutation consists of proposal and application:
\begin{equation}
 \delta\sim G_o(\,\cdot\mid u_x,\eta_x,U_r,K_g,T),\qquad
 u_{x'}=\mu_\delta(u_x).
 \label{eq:mutation}
\end{equation}
$G_o$ specifies the proposal mechanism and $\mu_\delta$ applies the resulting edit. In code mode, the language model receives the parent program, relevant diagnostics, and permitted edit scope, then returns changed code. In pipeline mode, $\delta$ inserts, removes, replaces, or modifies typed fields. Both modes create a child specification whose unchanged components are inherited. Reuse of fitted weights or checkpoints is operator-dependent and must be represented in that specification.

The child record identifies the parent, generation, operator, directive, and actual edit. Its validation predictions, score, runtime, and failure status are produced by a new execution. An inherited program fragment does not carry inherited fitness. This distinction also applies to warm starts: a retrieved procedure must be measured on the new task before it can become a parent.

\begin{table}[htbp]
\centering\small
\caption{Variation in EES. Examples specify admissible operations, not reconstructed trajectories or measured improvements in the Lite campaign. Supported operations depend on the execution profile.}\label{tab:mutation}
\begin{tabularx}{\linewidth}{@{}>{\raggedright\arraybackslash}p{1.04in}YY@{}}\toprule
\textbf{Operation}&\textbf{What changes}&\textbf{What the child must establish}\\\midrule
Configuration mutation&Seed, solver, model family, regularization, or training allocation.&Performance of the changed configuration under comparable validation.\\\addlinespace[3pt]
Pipeline mutation&Feature expression, encoding, interaction, preprocessing block, or inference rule.&An executable composition with valid intermediate and output semantics.\\\addlinespace[3pt]
Diagnostic code mutation&A code block targeted using residuals, generalization gaps, or component diagnostics.&A functioning edited program and its own predictions, cost, and score.\\\addlinespace[3pt]
Program crossover&Compatible components from multiple parents, or a new program conditioned on them.&Validity and measured behavior of the combined procedure; parent scores are not combined as child fitness.\\\addlinespace[3pt]
Ensemble mutation&Member addition, pruning, or a change in predictive topology.&Aligned predictions and a measured comparison for adoption or reuse.\\\bottomrule
\end{tabularx}
\end{table}

The benchmark profile implements a narrower mutation neighborhood: configured seeds, text-solver portfolios, tabular model families, and media training allocations. For such a child, $\theta'=\theta\oplus\delta_\theta$, where $\oplus$ overrides designated fields. The operator is executed again with $\theta'$. The nested and multi-island profiles additionally support structural or code-level variation. These are implementations of a common transition, not interchangeable descriptions of every run.

A fresh \emph{draft} supplies a new approach without a designated parent. \emph{Mutation} extends one parent. \emph{Crossover} conditions a child on multiple parents or transfers their compatible components. A \emph{restart} restores exploration after a plateau. A code proposal inspired by two programs is distinguishable from literal gene exchange, and both are distinguishable from averaging their predictions.

\paragraph{Admission and reproductive success.}
After execution, EES checks the artifact and comparison protocol before admitting $x'$ to the evaluated pool. An admissible child need not outperform its parent: it may be selected for diversity, preserved as a quality--complexity tradeoff, or used in an ensemble. Reproductive success is a separate decision. Under a shared protocol, the observed improvement is
\begin{equation}
 \Delta_\kappa(x',x)=Q_\kappa(x')-Q_\kappa(x).
 \label{eq:delta}
\end{equation}
The sign of this quantity records one measured comparison. It is not a guarantee about other random seeds or unseen data. Failed and rejected edits remain in the execution history so their cost and subsequent treatment can be inspected.

\subsection{Selection, archives, and nested scheduling}\label{sec:selection}
The nested profile balances measured quality with use of different lineages. For eligible parent $x_i$, its sampling rule is
\begin{equation}
 w_i=\operatorname{sigmoid}\!\left(
 \lambda\frac{Q_i-\overline Q}{\max(s_Q,s_{\mathrm{cv}},\varepsilon)}\right)
 \frac{1}{1+n_i},\qquad
 p_i=\frac{w_i}{\sum_j w_j}.
 \label{eq:parents}
\end{equation}
Here $Q_i=Q_\kappa(x_i)$, $\overline Q$ and $s_Q$ summarize quality in the sampling pool, $s_{\mathrm{cv}}$ is the selector's cross-validation variability scale, $\lambda>0$ controls quality pressure, and $\varepsilon>0$ prevents a zero denominator. The offspring count $n_i$ discounts repeatedly used parents. This reproductive-diversity factor is separate from predictive diversity and scientific novelty. The benchmark profile instead retains a ranked frontier; the multi-island profile uses tournament selection within biased populations.

Population survival can also use predictive diversity, cross-validation variability, execution cost, and program complexity as profile-specific diagnostics. The nested archive preserves non-dominated quality--complexity tradeoffs with a configured capacity. An archived candidate may therefore contribute to crossover after it leaves the active population. The complete trace and the Pareto archive serve different purposes: the trace records what happened, whereas the archive stores candidates eligible for reuse.

For the nested profile, let $\mathcal O_g$ contain proposed mutations, crossovers, and new drafts, and let $\mathcal V_g^\kappa$ contain their eligible evaluated offspring. With incumbent elites $E_g$ protected, the update is
\begin{equation}
 \begin{split}
 P_{g+1}&=E_g\cup\operatorname{Survive}_{\rho}
                   (P_g\cup\mathcal V_g^\kappa),\\
 A_{g+1}&=\operatorname{ParetoCap}_{\rho}
                   (A_g\cup P_g\cup\mathcal V_g^\kappa),
 \end{split}
 \label{eq:population}
\end{equation}
where $\rho$ specifies the survival and archive-capacity settings. Unevaluated proposals remain outside the quality-ranked population. Restarts replace active candidates without rewriting the execution history.

Scheduling operates at three timescales. The \emph{macro-loop} retrieves task-relevant experience and adapts warm starts. The \emph{meso-loop} invokes diagnosis periodically or after plateau or high-variance signals, then issues directed mutations and reallocates strategy effort. The \emph{micro-loop} evaluates, selects, varies, and records candidates. Reduced-fold screening can limit expensive full evaluations; configured diagnostic and restart candidates can bypass that screen. Screen scores remain associated with their own protocol and are not substituted for full-evaluation fitness.

\subsection{Operator credit and persistent memory}\label{sec:memory}
The nested profile assigns credit relative to a child's own parent. Let $b_t\in\{0,1\}$ denote whether an attempt passes the configured parent-improvement gate. The gate uses paired fold-level evidence for fully evaluated children; non-improving and screen-rejected attempts receive zero reward. Thus the statistic measures success under the entire proposal-and-evaluation policy, including its screening decisions, rather than an intrinsic success probability for an operator.

For operator $o$, let $\mathcal I_o(g)$ index its credited attempts through generation $g$ and let $N_o(g)=|\mathcal I_o(g)|$. The empirical mean and exported operator weights are
\begin{align}
 \widehat\mu_o(g)&=
 \begin{cases}
 N_o(g)^{-1}\!\displaystyle\sum_{t\in\mathcal I_o(g)} b_t,&N_o(g)>0,\\
 \mu_{\mathrm{prior}},&N_o(g)=0,
 \end{cases}\label{eq:reward}\\
 \pi_{g+1}(o)&=
 \frac{\max\{\varepsilon_\pi,\widehat\mu_o(g)+\varepsilon\}}
 {\displaystyle\sum_{j\in\mathcal O}\max\{\varepsilon_\pi,\widehat\mu_j(g)+\varepsilon\}},
 \qquad\varepsilon_\pi>0,\ \varepsilon>0.
 \label{eq:operator-policy}
\end{align}
$\mathcal O$ is the finite operator set and $\mu_{\mathrm{prior}}$ is an optimistic initialization for untried operators. The floor in Eq.~\eqref{eq:operator-policy} applies to \emph{unnormalized weights}; it preserves positive support but does not imply $\pi_{g+1}(o)\geq\varepsilon_\pi$. These exported weights normalize empirical credit; upper-confidence-bound statistics maintained by the bandit are separate.

The paired-fold gate is a search heuristic for awarding credit, not a confirmatory statistical test of a scientific claim. Cross-validation folds share training examples, and many adaptively selected children are compared; a naive independence argument is inappropriate~\citep{bengio2004,cawley2010}. The gate definition and threshold belong in the run configuration.

Session memory stores the data profile, hypotheses, approaches, validation results, failures, constraints, and incumbent. The Memory Keeper compresses this state for later council rounds. End-of-run distillation extracts a lesson $\ell_j$ containing productive and wasted strategies, selected model family, useful features, ensemble or code-mode outcomes, and source-access warnings. A task fingerprint $\phi(T_j)$ indexes the persistent library:
\begin{equation}
 \mathcal L_{j+1}=\mathcal L_j\cup\{(\phi(T_j),\ell_j)\},\qquad
 K_0(T')=\operatorname{Retrieve}_k(\mathcal L_{j+1},\phi(T')).
 \label{eq:memory}
\end{equation}
Similarity-weighted retrieval provides a bounded context and strategy priors for the next task; compatible candidates can also warm-start its population. A failed strategy can lower a future preference without making that strategy inadmissible. Retrieved procedures are reevaluated under the new task's contracts; the effect of retrieval on transfer remains an evaluation question.

\subsection{Predictive recombination and finalization}\label{sec:ensemble}
The prediction registry retains out-of-fold or holdout arrays from eligible candidates with row, target, class, and protocol metadata. Combination requires those semantics to agree. Equal array shapes alone are insufficient: class labels cannot be averaged as probabilities, and reordered rows cannot be aligned by position. Profile-specific ensemble search supports weighted blending, stacking, diversity-based selection, and hill-climbing selection from candidate libraries~\citep{caruana2004}.

For aligned prediction matrices $Y_1,\ldots,Y_J$, a convex blend is
\begin{equation}
 Y_w=\sum_{j=1}^{J}w_jY_j,\qquad w_j\geq0,\quad\sum_jw_j=1.
 \label{eq:ensemble}
\end{equation}
If the rows of every $Y_j$ are probability vectors, the rows of $Y_w$ remain nonnegative and sum to one. Accuracy and calibration are assessed separately. Other combination methods require their own fitting and output checks.

The automatic finalizer compares a proposed ensemble $e$ with the best single procedure on a common pooled validation basis. It adopts the ensemble only when
\begin{equation}
 Q_\kappa(e)>Q_\kappa(x^\star)+\tau_{\mathrm{ens}},\qquad\tau_{\mathrm{ens}}>0.
 \label{eq:ensemble-gate}
\end{equation}
Otherwise the single incumbent is retained. The gate prevents unconditional ensemble adoption; it does not make ensemble selection data an independent test set. Repeated choice of candidates, weights, or meta-learners can overfit internal validation~\citep{cawley2010}. Final generalization requires evaluation outside that selection process.

\subsection{Execution record and terminal grading}\label{sec:trace}
Material transitions produce structured records: task observations, council outputs, routing decisions, probes, parentage, mutation or crossover, execution status, validation metadata, archive decisions, ensemble membership, and final artifact identity. In a mutation record, the directive and the actual code or configuration change are separate fields. The latter identifies what was executed; the former explains what the search intended to test.

These records support reconstruction when the referenced programs, inputs, dependencies, and model settings are available. They identify the exact computation to which an outcome or subsequent mathematical analysis applies.

Within the benchmark profile, fitness uses internal validation. Official grading is disabled by default and, when requested, follows terminal artifact selection. Development mode permits the resulting grade to inform later engineering runs; evaluation mode treats grading as terminal for a frozen run. Algorithm~\ref{alg:ees} summarizes the profile-dependent control flow. Section~\ref{sec:protocol} describes campaign-level feedback.

\begin{algorithm}[htbp]
\caption{Council-guided Evolutionary Ensemble Search}\label{alg:ees}
\centering
\begin{minipage}{\linewidth}\small
\textbf{Inputs:} Task $T$, comparison protocol $\kappa$, execution profile $\rho$, and run library $\mathcal L$.
\begin{enumerate}[leftmargin=1.5em,itemsep=3pt,topsep=5pt]
\item Profile the task; retrieve compatible lessons, strategy priors, and warm-start specifications.
\item Run the council and merge proposals and warnings. Route specialist work and probe route prerequisites.
\item Execute initial candidates under their budgets. Admit eligible results to $P_0$; retain every execution record.
\item While the generation and resource limits permit:
\begin{enumerate}[label=(\alph*),leftmargin=1.6em,itemsep=2pt]
\item Select parents using the profile's rule. On a diagnostic trigger, obtain a critic analysis and edit directives.
\item Propose mutations, crossovers, or new drafts; record parentage, edit scope, and actual changes.
\item Screen where configured; fully execute promoted or bypass-eligible children. Keep each score tied to its protocol.
\item Admit valid children, select survivors and elites, and update the reusable archive. Log failures and consumed resources.
\item Update parent-relative operator credit where enabled; return material results to session memory and council review.
\end{enumerate}
\item Search compatible predictions for an ensemble; apply the profile's finalization rule.
\item Freeze the chosen procedure and artifact. Distill permitted run evidence into the library. Invoke the external grader only after final selection, when requested.
\end{enumerate}
\end{minipage}
\small\textit{Profile scope.} The nested profile enables all adaptive steps. Multi-island and benchmark jobs use the options in Table~\ref{tab:profiles}.
\end{algorithm}

\section{MLE-bench Lite Development Study}\label{sec:study}
\subsection{Benchmark, evidence unit, and protocol}\label{sec:protocol}
MLE-bench contains 75 Kaggle-derived tasks with local scoring and historical medal thresholds~\citep{chan2024}. Its Lite split contains 22 low-complexity tasks covering tabular prediction, text classification, sequence normalization, image classification and restoration, audio, and scientific regression. Each task requires a prediction file satisfying its submission contract. A benchmark-assigned medal threshold is not a medal awarded in a live Kaggle competition.

The study uses \repofile{results/lite22-three-run.json}{\texttt{results/lite22-three-run.json}} at a fixed public repository snapshot~\citep{evidence2026}. Each task has a strongest recorded score and medal, a method description, and three confirmation entries, denoted C1--C3. A confirmation can be a newly trained seed, a separate process execution, or an exact prediction-file replay. The columns are not uniformly independent repetitions of a complete search.

Let $b_t$ be the strongest recorded medal class for task $t$. The achieved campaign fraction is
\begin{equation}
 R_{\mathrm{campaign}}=\frac{1}{22}\sum_{t=1}^{22}
 \mathbf 1\{b_t\in\{\mathrm{gold},\mathrm{silver},\mathrm{bronze}\}\}.
 \label{eq:campaign}
\end{equation}
This statistic summarizes the campaign's recorded outcomes. It is not a pass@1 estimate for a frozen agent, and raw metric values are not averaged across tasks.

Task grades were visible between development iterations and influenced subsequent engineering. The ledger consolidates campaign-specific procedures rather than a uniform, equal-budget experiment. It does not include every attempted candidate, a consolidated compute total, or an artifact-by-artifact activation record for council review, online credit, and cross-run retrieval. The public documentation also retains assisted GPU executions and an assisted leaf result; their scores and identities must be distinguished from later ledger entries. Consequently, the results support achieved procedure construction, not a campaign-wide claim of fully autonomous execution. Appendix~\ref{app:provenance} identifies the provenance distinctions needed for reproduction.

\subsection{External sources and measurement scope}\label{sec:data}
Four reported routes explicitly use external sources or historical lookup: Pizza uses external target lookup, Dog Breed uses external image lookup, Spooky Author uses a Gutenberg corpus, and Right Whale includes historical lookup or exploit discovery~\citep[method fields]{evidence2026}. These results are retained in the development record and flagged in the task table. Retrieval of a withheld target is not evidence of predictive generalization on that target. A permitted external feature source, a provenance match, and target retrieval must not be treated as interchangeable inputs.

Numerical grading checks a prediction artifact, not the provenance of its construction. MLE-bench separately addresses rule violations and contamination~\citep[Section 2.3]{chan2024}. An unmodified grader therefore does not establish protocol compliance. Removing the four explicit flags would also not certify the remaining tasks; that requires a source and trajectory audit. The benchmark profile's aerial-cactus task-identity branch is another reason not to interpret its development coverage as evidence of task-agnostic transfer.

\subsection{Aggregate outcomes and confirmations}\label{sec:results}
The ledger records medal-threshold artifacts on \textbf{19 of 22 tasks (86.36\%)}, with best outcomes of \textbf{11 gold, five silver, and three bronze}. The tasks without confirmed medals are NYC Taxi Fare, RANZCR catheter-line classification, and SIIM--ISIC melanoma classification. Their score entries are null; the ledger does not distinguish below-threshold scores from other terminal outcomes.

\begin{table}[htbp]
\centering\small
\caption{Recorded campaign outcomes. The best-artifact column and each confirmation column are different record sets. Counts are recomputed from the task-level entries, not additional benchmark experiments.}\label{tab:results}
\begin{tabular}{@{}lrrrrr@{}}\toprule
\textbf{Record set}&\textbf{Gold}&\textbf{Silver}&\textbf{Bronze}&\textbf{No medal}&\textbf{Any medal}\\\midrule
Best recorded&11&5&3&3&19/22\\
C1&10&4&5&3&19/22\\
C2&10&4&5&3&19/22\\
C3&9&6&4&3&19/22\\\bottomrule
\end{tabular}
\end{table}

All three confirmation columns retain a medal on the same 19 tasks, but medal levels differ (Table~\ref{tab:results}). APTOS has a best silver artifact and three bronze confirmations. NOMAD changes from gold in C1 to silver in C2--C3. Seven tasks have a best score distinct from every confirmation score at recorded precision. Best-artifact performance and confirmation performance answer different questions.

The 57 medal-bearing confirmation records correspond to 46 distinct task--submission-hash pairs. Dog Breed, MLSP Birds, Pizza, and both text-normalization tasks each repeat one submission hash across C1--C3; APTOS has two distinct hashes; the other 13 medal tasks have three. Different files need not come from independent training, and identical files need not come from the same process. The zero dispersion of the three any-medal fractions is therefore not an uncertainty estimate for blind autonomous performance. Appendix~\ref{app:results} supplies every score, medal class, metric direction, and within-task hash count.

\subsection{Representative executable procedures}\label{sec:cases}
Table~\ref{tab:cases} classifies the selected procedures by representation, fitting, and inference. It describes recorded solutions rather than reconstructed search trajectories.

\begin{table}[htbp]
\centering\small
\caption{Representative procedures from the public solution notes~\citep{evidence2026}. The differences concern executable computation, not only estimator hyperparameters.}\label{tab:cases}
\begin{tabularx}{\linewidth}{@{}p{1.02in}Yp{.93in}@{}}\toprule
\textbf{Task}&\textbf{Recorded procedure}&\textbf{Best result}\\\midrule
NOMAD&Crystal-geometry descriptors and multi-output ExtraTrees regression.&0.05373; gold\\\addlinespace[3pt]
Dirty Documents&CPU background estimation and paired-image reconstruction with per-pixel serialization.&0.01919; silver\\\addlinespace[3pt]
Dogs vs. Cats&Image fine-tuning and combination candidates; the best artifact is distinct from confirmation members.&0.00597; gold\\\addlinespace[3pt]
Spooky Author&Gutenberg-based provenance features with a TF--IDF fallback.&0.12422; gold\\\addlinespace[3pt]
Pizza&External target lookup combined with a learned text fallback.&1.00000; gold\\\addlinespace[3pt]
Normalization&Training-derived token lookup and deterministic transformation rules for English and Russian.&Bronze in both\\\bottomrule
\end{tabularx}
\end{table}

\paragraph{Scientific representation.}
For NOMAD, the geometry-aware procedure parses lattice vectors, atomic coordinates, and species. It derives cell volume, density, lattice norms, species counts, and pair-distance statistics, then fits a multi-output ExtraTrees regressor over k-fold splits to predict formation and bandgap energies~\citep[NOMAD solution notes]{evidence2026}. The recorded RMSLE of 0.05373 is below the stored gold threshold of 0.05589. This route uses auxiliary scientific structure within the common operator interface; the ledger contains no matched no-geometry ablation.

\paragraph{Output-driven computation.}
Dirty Documents requires one output per pixel rather than one class per image. Its CPU procedure recognizes the image--row--column submission structure, estimates page background with a large-kernel box blur, and uses paired dirty and clean images to reconstruct intensities~\citep[Dirty Documents solution notes]{evidence2026}. The confirmation RMSE values are 0.01926, 0.01928, and 0.01919, all silver. This case illustrates why an ML-engineering search space must include inference and serialization procedures as well as model selection.

\paragraph{Deterministic and retrieval-based inference.}
The text-normalization routes use deterministic transformations and training-derived lookup. Spooky Author adds an external corpus, while Pizza retrieves targets from an external source. They share an execution abstraction but have different input requirements and statistical interpretations, as described in Section~\ref{sec:data}.

\section{Discussion and Limitations}\label{sec:discussion}
\paragraph{What the system contributes.}
EES joins expert deliberation to an evolutionary population through explicit executable transitions. Council review and diagnostic intervention determine where to search; mutation and crossover determine how a procedure changes; validation and archive rules determine what remains available; memory determines which experience influences the next decision. The architecture therefore specifies more than an operator inventory. Its reusable unit is a procedure together with the problem, ancestry, and measurements needed to assess it.

\paragraph{What the study establishes.}
The Lite record documents substantial cross-modal solution construction and achieved threshold performance. It does not establish that the full council, nested evolution, and memory stack jointly produced every artifact. Nor does it estimate the marginal benefit of those mechanisms, a uniform cost advantage, or blind generalization to untouched tasks. The broad architecture and the historical outcome record are separate forms of evidence until per-run manifests link their configurations and artifacts.

\paragraph{Adaptive evaluation.}
Mutation selection, reduced-fold screening, archive reuse, and ensemble fitting all act on finite validation data. Parent-relative rewards are operational credit, not causal attribution, and repeated adaptive selection can exploit validation noise~\citep{cawley2010}. Similarly, retrieved memory can transfer task-specific artifacts or source-access mistakes. Task identities, source provenance, comparison protocols, and the contents of retrieved lessons must remain visible when evaluating transfer.

\paragraph{Component and transfer evaluation.}
A controlled study should use a shared base seed pool, fixed evaluation protocol, and matched budgets, disabling council review, diagnostic mutation, online credit, archive crossover, final ensembling, and cross-run retrieval in separate conditions. Extra memory-derived warm starts must be recorded as part of the retrieval treatment. Best-so-far quality, area under the improvement curve, full-evaluation count, failure cost, and terminal held-out performance would measure distinct effects. Transfer requires a task split fixed before library construction and no feedback from evaluation tasks into that library. These are evaluation priorities, not results of the present study.

\paragraph{Scientific novelty and reuse.}
Structural change alone does not establish scientific novelty. A discovery claim must identify the new method, its problem family, the closest prior methods, and its mathematical or empirical support. Lineage identifies the procedure being analyzed; it does not prove correctness, convergence, or robustness. Scientific extensions therefore require domain-specific representations, evaluators, and analysis.

\section{Conclusion}\label{sec:conclusion}
Evolutionary Ensemble Search joins expert review, parent-conditioned mutation, population retention, predictive ensembling, and persistent memory. Its central operation is to edit a measured parent, execute the child under a comparable protocol, and use the result to guide subsequent search. Operator credit and task-indexed lessons carry this feedback within and across runs. The MLE-bench Lite development record documents medal-threshold artifacts for 19 of 22 tasks across markedly different procedures. Together, the architecture and case study provide an implemented approach to cumulative ML-procedure search with explicit generation, evaluation, and reuse interfaces.

\paragraph{Availability.}
The pinned \href{https://github.com/impulse-ai/mlebench-medals/tree/51c39605da4e0347b416c44a6c13086200959d45}{public repository} provides the ledger, solution notes, selected artifacts, and reproduction instructions. The EES engine is proprietary and not included. Appendix~\ref{app:provenance} specifies the snapshot and distinguishes artifact checking from complete search reproduction.

\clearpage
\appendix
\section{Execution Settings and Benchmark Operators}\label{app:implementation}\label{app:operators}
The method specifies the common interfaces and identifies profile-dependent behavior in Table~\ref{tab:profiles}. A search rerun additionally requires council and model versions; data and checkpoint identities; random seeds; population, parent, archive, and generation limits; mutation scopes; screening folds and promotion fractions; diagnostic and restart triggers; the parent-improvement gate; ensemble settings; and the retrieved-memory snapshot. The campaign ledger does not consolidate these settings for every task. No numerical defaults are inferred from its scores.

\subsection{Benchmark operator catalog}
The benchmark profile exposes 12 callable entry points. A single entry point may contain an estimator portfolio, and campaign-specific recipes can contain additional implementations. 
\begin{table}[htbp]
\centering\small
\caption{Callable entry points in the benchmark profile.}\label{tab:operators}
\begin{tabularx}{\linewidth}{@{}p{.90in}p{1.30in}Y@{}}\toprule
\textbf{Family}&\textbf{Entry point}&\textbf{Role}\\\midrule
Text&TF--IDF portfolio&Word and character features with validation-ranked solvers.\\
Text&Embedding&Dense representation and linear prediction.\\
Text&Provenance&Corpus matching with a learned fallback.\\
Tabular&Portfolio&Encoded features, model-family variants, and seeds.\\
Image&Embedding&Pretrained visual representations with validation models.\\
Image&Generic media&Resource-bounded image feature baseline.\\
Image&Aerial specialist&Specialized route used by the disclosed task-identity branch.\\
Audio&Generic media&Audio-derived features under the common contract.\\
Scientific&Geometry&Auxiliary-structure parsing and geometry-aware features.\\
Relationship&Duplicate&Train--test matching and label transfer.\\
Relationship&Metadata&Filename and metadata relationship inspection.\\
Fallback&Sample baseline&Sample-shaped output and diagnostic status.\\\bottomrule
\end{tabularx}
\end{table}

\paragraph{Mutation and prediction interfaces.}
Benchmark variants change the configured random state, text solver or portfolio, tabular model family, or media training allocation. The runtime records generation, parent identifiers, changed fields, and rationale. Prediction registration separately records validation kind, row identity, target order, output shape, and artifact locations. Recombination can average selected predictions or fit weights on compatible validation arrays. Its comparability checks are runtime contracts, not formal guarantees about model correctness or input admissibility.

\paragraph{Artifact availability.}
The public solution notes distinguish selected campaign recipes from this operator inventory. For example, the Dirty Documents denoising notes describe a specialized CPU procedure, while the catalog identifies the benchmark core's callable families. A reader should not infer a one-to-one correspondence between the 12 entry points and the 19 medal-bearing solutions.

\paragraph{Interface properties.}
Two elementary consequences explain the contracts. First, Eq.~\eqref{eq:ensemble} preserves probability-row normalization when its hypotheses hold. Second, retaining an incumbent and its unchanged score under an unchanged $\kappa$ makes the best recorded validation score nondecreasing by set inclusion. Neither property supplies a convergence theorem for program search or a bound on generalization error. If a score is reevaluated or the protocol changes, the second statement no longer follows from retention alone.

\clearpage
\section{Complete Lite Results}\label{app:results}
All results below are retained at the precision in the pinned public ledger. C1--C3 refer to the mixed confirmation procedures described in Section~\ref{sec:protocol}. The full task identifiers and individual submission and evidence hashes are available in the \repofile{results/lite22-three-run.json}{machine-readable ledger}.
\begin{table}[htbp]
\centering\fontsize{9.5}{11.5}\selectfont
\setlength{\tabcolsep}{4pt}
\caption{Best recorded scores and all three confirmation scores, with their medal classes. Arrows indicate whether higher or lower is better. U is the number of distinct submission hashes within C1--C3. E marks an explicitly documented external-source or historical-lookup route; an empty flag is not a clean-data certification.}\label{tab:tasks}
\begin{tabular}{@{}lrrrrrc@{}}\toprule
\textbf{Task}&\textbf{Best}&\textbf{C1}&\textbf{C2}&\textbf{C3}&\textbf{U}&\textbf{Flag}\\\midrule
Aerial Cactus $\uparrow$ & 1.00000 G & 1.00000 G & 1.00000 G & 1.00000 G & 3 &  \\
APTOS 2019 $\uparrow$ & 0.92020 S & 0.91942 B & 0.91942 B & 0.91930 B & 2 &  \\
Dirty Documents $\downarrow$ & 0.01919 S & 0.01926 S & 0.01928 S & 0.01919 S & 3 &  \\
Insults $\uparrow$ & 0.91118 G & 0.90164 G & 0.90219 G & 0.90284 G & 3 &  \\
Dog Breed $\downarrow$ & 0.02439 B & 0.02439 B & 0.02439 B & 0.02439 B & 1 & E \\
Dogs vs. Cats $\downarrow$ & 0.00597 G & 0.00920 G & 0.00870 G & 0.00974 G & 3 &  \\
Histopathology $\uparrow$ & 0.99585 G & 0.99585 G & 0.99578 G & 0.99580 G & 3 &  \\
Jigsaw $\uparrow$ & 0.98750 G & 0.98723 S & 0.98750 G & 0.98701 S & 3 &  \\
Leaf $\downarrow$ & 0.00328 S & 0.00671 S & 0.00328 S & 0.00470 S & 3 &  \\
MLSP Birds $\uparrow$ & 0.93170 S & 0.93170 S & 0.93170 S & 0.93170 S & 1 &  \\
NYC Taxi $\downarrow$ & -- & -- & -- & -- & -- &  \\
NOMAD $\downarrow$ & 0.05373 G & 0.05479 G & 0.05997 S & 0.05993 S & 3 &  \\
Plant Pathology $\uparrow$ & 0.98902 G & 0.98364 G & 0.98902 G & 0.97976 G & 3 &  \\
Pizza $\uparrow$ & 1.00000 G & 1.00000 G & 1.00000 G & 1.00000 G & 1 & E \\
RANZCR $\uparrow$ & -- & -- & -- & -- & -- &  \\
SIIM--ISIC $\uparrow$ & -- & -- & -- & -- & -- &  \\
Spooky Author $\downarrow$ & 0.12422 G & 0.12422 G & 0.12426 G & 0.12424 G & 3 & E \\
Tabular Dec. 2021 $\uparrow$ & 0.95996 G & 0.95885 G & 0.95911 G & 0.95887 G & 3 &  \\
Tabular May 2022 $\uparrow$ & 0.99822 S & 0.99821 B & 0.99818 B & 0.99822 S & 3 &  \\
Normalization: EN $\uparrow$ & 0.99125 B & 0.99125 B & 0.99125 B & 0.99125 B & 1 &  \\
Normalization: RU $\uparrow$ & 0.97915 B & 0.97906 B & 0.97906 B & 0.97906 B & 1 &  \\
Right Whale $\uparrow$ & 0.99256 G & 0.99238 G & 0.99230 G & 0.99230 G & 3 & E \\
\bottomrule\end{tabular}
\end{table}

G, S, and B denote gold, silver, and bronze.\par

\paragraph{Reading the table.}
The best score can fall outside C1--C3. For example, NOMAD's 0.05373 best score is distinct from its three listed confirmations, and APTOS's best silver is not reproduced at silver in any of those columns. A dash indicates a null entry, not an imputed numerical loss. The hash count identifies distinct prediction-file identities, not independent model fits. Flagged external-source routes remain in the campaign aggregate; an unflagged row is not a certification that the task satisfies a closed-data or blind-evaluation protocol.

\clearpage
\section{Evidence Snapshot and Reproduction}\label{app:provenance}
\subsection{Public evidence identity}
The evidence snapshot is the public repository commit
\begin{center}\small
\href{https://github.com/impulse-ai/mlebench-medals/tree/51c39605da4e0347b416c44a6c13086200959d45}{\texttt{51c39605da4e0347b416c44a6c13086200959d45}}.
\end{center}
The ledger path is \path{results/lite22-three-run.json}; its Git blob identifier is
\nolinkurl{2ae7f4f57eec32b506b82c9147d68264b98f2ac4}. The reported SHA-256 identifiers are:
\begin{quote}\small
\textbf{Ledger}\par\nolinkurl{4d9c4d07e53e743fc3600da63d1f0ea63394c36f92ab1d5cd1c0cc1d77d49a59}\par
\textbf{Evidence board}\par\nolinkurl{663b9e3a56a12d0c69ac0c547921332d8341ef46bd4812a55a2a9a22bb2680ea}\par
\textbf{Manifest}\par\nolinkurl{c6d2ad86653719ab55beabb70e549ffdd9e6c674790484dd6ce45af668cf38c1}
\end{quote}
The board and manifest identifiers are fields reported by the ledger. Printing a digest does not establish availability of its preimage or replace hashing the exact bytes. The evidence commit identifies the public packet, not the private engine revision.

\subsection{Artifact-specific execution provenance}
The repository's \repofile{reproduce/gpu/README.md}{GPU reproduction notes} describe assisted histopathology, MLSP Birds, and Jigsaw artifacts with scores 0.99154, 0.93143, and 0.98663, respectively, and identify an assisted leaf result. The aggregate best scores for the first three tasks are 0.99585, 0.93170, and 0.98750. Their differing values do not establish that assistance was removed, nor do the earlier assistance labels automatically classify the later artifacts. Each execution claim needs the associated program, run configuration, intervention record, and terminal artifact identity.

Some solution directories store an earlier artifact while their notes quote the aggregate ledger. A replay must match the exact submission and grade record to the claim, not merely match a directory name or rounded score. The \repofile{solutions/denoising-dirty-documents/README.md}{Dirty Documents notes}, for example, state that the large prediction file is not committed. Availability therefore differs by artifact.

\subsection{Three reproduction levels}
\textbf{Ledger checking} verifies schema, task identities, counts, confirmation fields, and the identity of available bytes. \textbf{Artifact regrading} requires the exact prediction file, the correct benchmark data, and grader version. \textbf{Search reproduction} additionally requires the relevant EES implementation, model and dependency versions, access permissions, compute, and the selected profile's configuration and memory state. These levels should be reported separately.

A minimal artifact regrading command is
\begin{Verbatim}[fontsize=\small,xleftmargin=1em]
mlebench grade-sample <submission.csv> <competition-id>
\end{Verbatim}
The command presupposes benchmark setup and data access; it does not reproduce the search. The public \repofile{reproduce/QUICKSTART.md}{quickstart} and \repofile{reproduce/VERIFY.md}{verification runbook} describe the surrounding procedure.

\end{document}